\documentclass[letterpaper]{article} 
\usepackage{aaai2026}  
\usepackage{times}  
\usepackage{helvet}  
\usepackage{courier}  
\usepackage[hyphens]{url}  
\usepackage{graphicx} 
\usepackage{natbib}  
\usepackage{caption} 
\usepackage{algorithm}
\usepackage{algorithmic}
\usepackage{xcolor}
\usepackage{amsmath} 
\usepackage{amsfonts}
\usepackage{multirow}
\usepackage{booktabs}

\usepackage{newfloat}
\usepackage{listings}
\DeclareCaptionStyle{ruled}{labelfont=normalfont,labelsep=colon,strut=off} 
\floatstyle{ruled}
\newfloat{listing}{tb}{lst}{}
\floatname{listing}{Listing}
\title{SFedHIFI: Fire Rate-Based Heterogeneous Information Fusion for Spiking Federated Learning}
\author {
    Ran Tao\textsuperscript{\rm 1, \rm 2},
    Qiugang Zhan\textsuperscript{\rm 1, \rm 2},
    Shantian Yang\textsuperscript{\rm 1, \rm 2, \rm 3},
    Xiurui Xie\textsuperscript{\rm 4},
    Qi Tian\textsuperscript{\rm5, \rm 1},
    Guisong Liu\textsuperscript{\rm 1, \rm 2, \rm 3}\thanks{Corresponding authors.}
}
\affiliations {
    \textsuperscript{\rm 1}Complex Laboratory of New Finance and Economics, School of Computing and Artificial Intelligence, Southwestern University of Finance and Economics, China\\
    \textsuperscript{\rm 2}Engineering Research Center of Intelligent Finance, Ministry of Education, China\\
    \textsuperscript{\rm 3}Kash Institute of Electronics and Information Industry, China\\
    \textsuperscript{\rm 4}Laboratory of Intelligent Collaborative Computing, University of Electronic Science and Technology of China, China\\
    \textsuperscript{\rm 5}Huawei Inc., China\\
    rantaostd@gmail.com, \{zhanqg, yangst, gliu\}@swufe.edu.cn, xiexiurui@uestc.edu.cn,  tian.qi1@huawei.com
}

\begin{document}

\maketitle

\begin{abstract}
Spiking Federated Learning (SFL) has been widely studied with the energy efficiency of Spiking Neural Networks (SNNs). However, existing SFL methods require model homogeneity and assume all clients have sufficient computational resources, resulting in the exclusion of some resource-constrained clients. 
To address the prevalent system heterogeneity in real-world scenarios, enabling heterogeneous SFL systems that allow clients to adaptively deploy models of different scales based on their local resources is crucial. 
To this end, we introduce SFedHIFI, a novel \textbf{S}piking \textbf{Fed}erated Learning framework with Fire Rate-Based \textbf{H}eterogeneous \textbf{I}nformation \textbf{F}us\textbf{i}on.
Specifically, SFedHIFI employs channel-wise matrix decomposition to deploy SNN models of adaptive complexity on clients with heterogeneous resources. 
Building on this, the proposed heterogeneous information fusion module enables cross-scale aggregation among models of different widths, thereby enhancing the utilization of diverse local knowledge.
Extensive experiments on three public benchmarks demonstrate that SFedHIFI can effectively enable heterogeneous SFL, consistently outperforming all three baseline methods. Compared with ANN-based FL, it achieves significant energy savings with only a marginal trade-off in accuracy.
\end{abstract}


\section{Introduction}
\begin{figure*}[ht]
  \begin{center}
    \includegraphics[width=\textwidth]{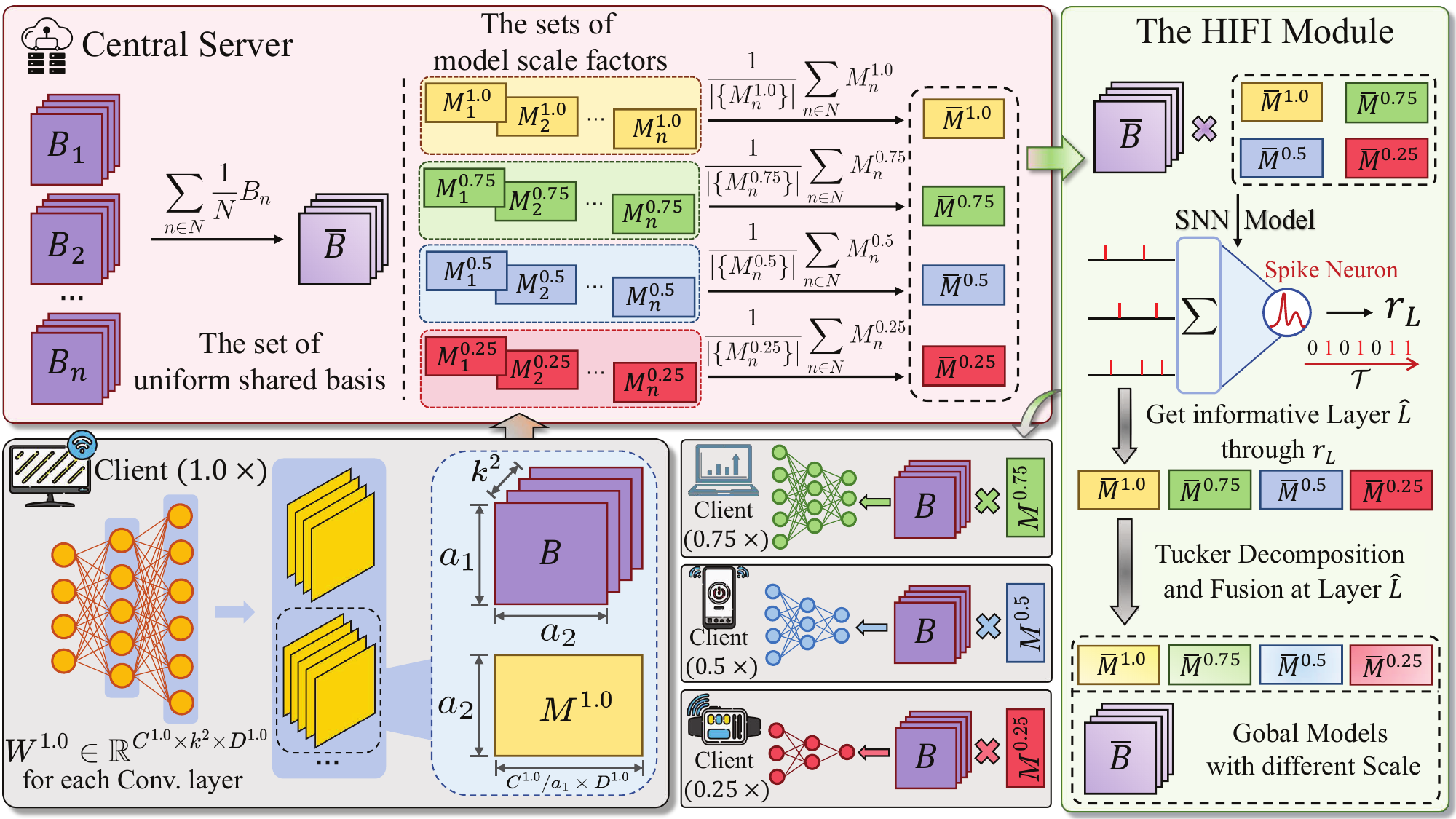}
  \end{center}
   \caption{Overview of SFedHIFI. First, local clients are clustered into the model scale set of computational resources $\mathbf{p}=\{0.25,0.5,075,1.0\}$. Then, the local $(\mathbf{p} \times)$ model gets the uniform shared basis $B$ and the model scale factor $M^{\mathbf{p}}$ through the Channel-wise Matrix Decomposition. In the central server, the uniform shared basis will be aggregated as the global $\overline{B}$, and the model scale factors will be aggregated into global $\{\overline{M}^{\mathbf{p}}\}$ corresponding to different model scales. The list of global model scale factors $\{\overline{M}^{\mathbf{p}}\}$ is then fed to the HIFI module, where the Firing Rate-Based Heterogeneous Information Fusion is applied to integrate knowledge across scales. This process enhances the representational capacity of global SNNs at all scales and improves their overall performance.}
   \label{fig:overview}
\end{figure*}

Spiking Federated Learning (SFL) is an emerging research paradigm that integrates Spiking Neural Networks (SNNs)~\cite{1_maass1997networks} with Federated Learning (FL)~\cite{sec1_FLsurvey_2021,sec1_FLsurvey_pami}, aiming to build FL systems tailored for resource-constrained edge environments. 
In particular, while FL serves as a general framework for privacy-preserving distributed learning, resource limitations of edge participants remain a critical challenge. 
Notably, SNNs employ an event-driven processing paradigm, enabling sparse and energy-efficient computation, especially when deployed on neuromorphic hardware, making them particularly suitable for resource-constrained environments~\cite{sec1_spikechip1,sec1_loihi,sec1_truenorth,sec1_Tianjic}.
Current SFL research primarily focuses on reducing the system power consumption by optimizing distributed training protocols and communication-efficient mechanisms~\cite{sec1_fedwithsnn_2020,sec1_fedwithsnn_fedsnn,sec1_fedwithsnn_nature,sec1_fedwithsnn_xie}, demonstrating significant advancements in various common tasks (e.g., image classification).

Despite considerable progress in existing SFL, these approaches typically impose a strict requirement for architectural homogeneity between client models and the global one. 
This uniformity constraint proves overly restrictive for practical deployment scenarios. 
In contrast, participants typically utilize diverse edge devices capable of gathering valuable data, which are constrained by computational limitations and exhibit significant heterogeneity in resource budgets.
This widespread system heterogeneity has emerged as a major bottleneck for the practical deployment of SFL applications in real-world scenarios.

To address system heterogeneity in FL, ANN-based approaches typically either exclude resource-constrained edge devices and their distinctive data knowledge~\cite{sec1_clientselect2020,sec1_clientselect2023} or reduce resource demands through model compression techniques~\cite{sec1_modelcompress2021,sec1_modelcompress2022}, such as decreasing model depth (the number of layers) or width (the number of hidden channels). Motivated by the above strategies and recent advances in SNN pruning~\cite{sec2_2019snnft,sec2_2023snnft}, we conducted preliminary explorations into a heterogeneous SFL system that combines channel pruning with conventional aggregation algorithms.
However, as demonstrated by our empirical results in Table~\ref{tab1:benmarkexp}, the inherently sparse feature representations in SNNs are particularly susceptible to information loss during model compression, leading to degraded performance under heterogeneous settings. 
Hence, developing a heterogeneous SFL framework based on channel pruning is a promising direction. It can effectively train and aggregate information across diverse SNNs, yet remains underexplored.

In this paper, we propose SFedHIFI, a novel \textbf{S}piking \textbf{Fed}erated Learning framework with Fire Rate-Based \textbf{H}eterogeneous \textbf{I}nformation \textbf{F}us\textbf{i}on.
SFedHIFI supports the configuration of complex global SNN models and introduces a dynamic channel pruning strategy to modify the width of local models for resource adaptation.
Specifically, we introduce a channel-wise matrix decomposition technique~\cite{sec1_2017decomptech1,sec1_2020decomptech2} to project the parameters of SNNs with varying widths onto a shared $B$ matrix representing the global channel space and a client-specific $M$ matrix encoding heterogeneous channel dimensions. 
In this framework, $B$ facilitates cross-scale aggregation from all models to capture broad knowledge, whereas $M$ enables scale-specific aggregation of heterogeneous information, ensuring stable model convergence. Following aggregation, the matrices $B$ and $M$ are distributed to the corresponding clients. 
This approach guarantees universal access to distributed knowledge while significantly enhancing the representational diversity and generalization capacity of the global model.

Furthermore, due to the sparse and binary nature of SNN representations, the shared $B$ matrix tends to capture limited global knowledge, while the client-specific matrices $M$ retain potentially valuable cross-scale heterogeneous information. 
To leverage this, we propose a novel heterogeneous information fusion (HIFI) module tailored for the client-specific $M$ matrices to improve heterogeneous aggregation.
Specifically, we analyze spiking neural dynamics to identify high-information-density layers within the $M$ matrices, then decouple these valuable parameters and fuse them across different model scales.

Experimental results demonstrate that SFedHIFI, leveraging the spiking neural dynamics to address the information sparsity challenge inherent in SNNs, significantly enhances the performance of heterogeneous aggregation. 
The main contributions of our studies are summarized as follows:
\begin{itemize}
    \item We propose SFedHIFI, a novel heterogeneous SFL framework that enables clients to adaptively deploy SNN models of varying widths via channel pruning, while effectively aggregating heterogeneous SNN information to improve global model performance across different scales.
    \item To address the challenge of severe information loss caused by SNN compression, we design a new HIFI module based on layer firing rates, which further exploits the intrinsic heterogeneous knowledge within SNNs and enhances the overall performance of SFedHIFI.
    \item Extensive experiments on four benchmark datasets demonstrate that SFedHIFI effectively mitigates information degradation in compressed SNNs, achieving an energy-efficient heterogeneous SFL paradigm.
\end{itemize}

\section{Preliminary and Related Work}

\subsection{The Leaky Integrate-and-Fire Model}
In SNNs, information is encoded as binary spike sequences through spiking neural models, typically the Leaky Integrate-and-Fire (LIF) neurons~\cite{sec2_lif_book_2014,sec2_lif2021}:
\begin{align}
    & H[t]=V[t-1]-\tau\left(V[t-1]-V_{\text{reset}}\right)+I[t], \\
    & S[t]=\Theta\left(V[t]-V_{\text{th}}\right), \\
    & V[t]=H[t]\cdot(1-S[t])+V_{\text{reset}}\cdot S[t],
\end{align}
where $\tau$ is the leaky factor, $V[t]$ represents the membrane potential of the neuron at time step $t$, $I[t]$ is the input for neuron. $S[t]$ denotes the output state of a neuron, taking a value of 1 or 0 to indicate whether it is firing a spike or not. $\Theta(\cdot)$ is the Heaviside function. When the membrane potential exceeds the threshold $V_{\text{th}}$, the neuron will trigger a spike and reset its membrane potential to $V_{\text{reset}}$.

\subsection{Spiking Federated Learning}
Emerging research is exploring Spiking Federated Learning (SFL) to alleviate pervasive resource constraints at the edge in FL. 
Seminal work~\cite{sec1_fedwithsnn_2020} pioneered on-device collaborative federated learning with SNNs. 
Tumpa et al. explored the integration of the Skipper technique~\cite{sec2_SFL-3_skipper_2022} for optimizing SNN training protocols, effectively constraining training memory across clients~\cite{sec2_SFL-4_2023}. 
For communication efficiency in SFL, HDSFL leveraged hint-layer distillation~\cite{sec1_fedwithsnn_xie}, and Nguyen et al. introduced a Top-k sparsification strategy~\cite{sec2_SFL_robust_2024}.
Other studies focus on improving the effectiveness of SFL. For example, AdaFedAsy aimed to mitigate the negative effects of stale updates from resource-constrained devices~\cite{sec2_SFL-5_2023}, while Yu et al. explored the integration of higher-accuracy ANNs into the SFL framework~\cite{sec2_SFL-HFL-csnn_2024}. 
Collectively, these advancements substantiate the effectiveness of SFL and highlight its inherent advantages in energy efficiency.

Nevertheless, current SFL studies universally assume identical model architectures, making them unsuitable for the more common system heterogeneity scenarios. 
In this paper, we fill this gap and demonstrate the effectiveness of heterogeneous SNNs in federated learning, thereby broadening the applicability of SFL.

\subsection{System-Heterogeneous Federated Learning}
To tackle system heterogeneity, SFL can draw inspiration from ANN-based FL studies. 
System heterogeneity primarily imposes resource constraints, which have been mitigated through accelerated convergence~\cite{sec2_sysh_ac_1}, model compression~\cite{sec2_sysh_modelcomp}, and sparsified training~\cite{sec2_sysh_sparsity}.
However, these methods cause significant information loss in SNNs due to sparse binary spikes. 
Moreover, FL based on knowledge distillation (KD) can support model heterogeneity, but its dependence on public datasets conflicts with privacy principles~\cite{sec2_kd}. 

Other works leverage model pruning to assign client-specific subnetworks of different sizes, enabling adaptive knowledge contribution~\cite{sec2_2021heterofl,sec1_HFL_fjord,sec1_HFL_rlanc,sec2_2024fedlps}.
Given the effectiveness of SNN pruning~\cite{sec2_2019snnft,sec2_2023snnft}, this provides valuable insight for heterogeneous SFL.
Yet, these methods restrict aggregation to shared channels, which benefits dense ANNs but discards essential information in sparse SNNs, degrading performance (see Table~\ref{tab2:ablationexp}).
In contrast, our method estimates feature information density from neuron firing rates to guide the aggregation of heterogeneous SNNs, effectively improving the global model.

\section{Method}
In this section, we first introduce the heterogeneous federated learning framework based on channel-wise matrix decomposition. Building upon this channel-pruned heterogeneous structure, we propose the HIFI module tailored for SNNs, which analyzes the firing rates across SNN layers to identify high-information-density layers within the client-specific $M$ matrices. These informative layers are then decoupled using Tucker decomposition to enable effective heterogeneous information fusion.
At last, we explain the local objective in SFedHIFI.

\subsection{Heterogeneous FL Framework Based on Channel-Wise Matrix Decomposition}
Considering prior studies that have demonstrated the effectiveness of channel pruning in SNNs, our work adopts a channel-wise matrix decomposition strategy for model pruning, enabling the construction of a heterogeneous SFL framework that supports cross-scale model aggregation.
Specifically, the heterogeneous FL framework comprises $N$ local clients with models of varying scales and a central server.
Every client $n \in\{1, \ldots, N\}$ has own local dataset.
At the start, clients are clustered into the model scale set of computational resources $\mathbf{p} = \{p_i|p_i \in (0,1], i<N\}$, according to their local computational resources. 
Next, the server assigns the uniform shared basis $B$ and the model scale factor $M^{\mathbf{p}}$ to each client according to the clustering results. 

For a client $n$, its local model width is $p_i\times$ full model, and the channel-wise matrix decomposition of the convolution weight parameter $\mathbf{W}_{n}^{p_i}$ can be described as,
\begin{equation}
    \begin{array}{cc}
    \label{eq:mdec4}
         & \mathbf{W}_{n}^{p_i} \approx B_n\cdot M_n^{p_i}, \\
         & \mathbf{W}_{n}^{p_i} \in \mathbb{R}^{C^{p_i} \times k^2 \times D^{p_i}}, \\
         &B_n \in \mathbb{R}^{k^2 \times a_1 \times a_2}, M_n^{p_i} \in \mathbb{R}^{a_2 \times \frac{C^{p_i}}{a_1} \times D^{p_i}},
    \end{array}
\end{equation}
where $C^{p_i}$ denotes the output channel with scale $p_i$, $D$ means the input channel, and $k$ is the convolution kernel size. $a_1$ and $a_2$ are pre-defined hyperparameters used to construct the unified shared basis $B$. 

Therefore, the shared basis $B$ and the scale factor $M^{p_i}$ are updated during every local training, then sent to the server for aggregation after local training. Specifically, the central server aggregates $B$ for all clients, and aggregates $M^{p_i}$ for all clients with scale $p_i$. The global $\overline{B}$ and $\overline{M}^{p_i}$ can be expressed as follows:
\begin{equation}
\label{eq:aggreB}
    \begin{array}{cl}
        \overline{B} &= \text{FA}(\{B_n\}_{n\in N}) \\
        &=\sum\limits_{n \in N} \frac{1}{N}B_n \in \mathbb{R}^{k^2 \times a_1 \times a_2},
    \end{array}
\end{equation}
\begin{equation}
\label{eq:aggreM}
    \begin{array}{cl}
        \overline{M}^{p_i} &= \text{FA}(\{M_n^{p}|p=p_i,n \in N\}) \\
        &=\frac{1}{|\{M_n^{p_i}\}|}\sum\limits_{n \in N} M_n^{p_i} \in \mathbb{R}^{a_2 \times \frac{C^{p_i}}{a_1} \times D^{p_i}},
    \end{array}
\end{equation}
where $B_n$ is the $B$ matrix of client $n$, and $\{M_n^{p_i}\}$ is the set of $M$ with scale $p_i$ from all clients. $\text{FA}(\cdot)$ means the FedAVG algorithm~\cite{sec3_fedavg}. 
After aggregation, the server would get the global $\overline{B}$ and $\{\overline{M}^\mathbf{p}\}$ list, and then distributes the set of global model weights $\{\overline{\mathbf{W}}^\mathbf{p}=\overline{B} \cdot \overline{M}^\mathbf{p}\}$ for every scale in set $\mathbf{p}$.

However, experimental results reveal that SNNs perform suboptimally under the heterogeneous FL framework based solely on channel-wise matrix decomposition (as evidenced by the performance of $\text{HIFI}_{w/o}$ in Table~\ref{tab2:ablationexp} compared to the Baseline methods in Table~\ref{tab1:benmarkexp}). To address this limitation and enhance the effectiveness of SNNs in heterogeneous settings, we further propose the core component of the SFedHIFI framework—the HIFI module.

\subsection{Firing Rate-Based Heterogeneous Information Fusion (HIFI) Module}
\begin{table*}[htbp]
\begin{center}
\resizebox{1.0\linewidth}{!}{
\begin{tabular}{cccccccccccccc}
\hline
\hline
\specialrule{0em}{0pt}{3pt}
 \multirow{2}{*}{\textbf{Dataset}} & \multirow{2}{*}{\textbf{Scale}} & \multicolumn{4}{c}{\textbf{IID}} & \multicolumn{4}{c}{\textbf{Non-IID} Dir(0.5)} & \multicolumn{4}{c}{\textbf{Non-IID} Dir(0.3)} \\
 \cmidrule(lr){3-6} \cmidrule(lr){7-10} \cmidrule(lr){11-14}
   &  & FedAVG & FedProx & FedNova & \textbf{SFedHIFI} & FedAVG & FedProx & FedNova & \textbf{SFedHIFI} & FedAVG & FedProx & FedNova & \textbf{SFedHIFI} \\
 \hline
 \specialrule{0em}{0pt}{2.5pt}
 \multirow{5}{*}{Fasion-MNIST} & $0.25\times$ & 90.50 & 89.10 & 88.48 & \textbf{91.46} & 85.19 & \textbf{89.21} & 88.59 & 88.70 & 83.35 & 86.24 & \textbf{87.74} & 87.27 \\
  & $0.5\times$ & 91.12 & 89.56 & 84.85 & \textbf{92.38} & 85.36 & 88.56 & 87.63 & \textbf{89.85} & 83.86 & \textbf{88.70} & 87.97 & 88.31 \\
  & $0.75\times$ & 91.47 & 89.68 & 73.70 & \textbf{92.03} & 88.18 & 89.69 & 89.82 & \textbf{90.41} & 86.14 & 88.91 & \textbf{89.31} & 88.99 \\
  & $1.0\times$ & 91.17 & 90.53 & 90.20 & \textbf{92.43} & 89.91 & \textbf{90.79} & 90.74 & 90.32 & 87.98 & 88.29 & 88.63 & \textbf{89.18} \\
  & Avg. & 91.06 & 89.97 & 84.31 & \textbf{92.08} & 87.16 & 89.56 & 89.20 & \textbf{89.82} & 85.33 & 88.04 & 88.41 & \textbf{88.43} \\
 \hline
 \specialrule{0em}{0pt}{2.5pt}
 \multirow{5}{*}{CIFAR-10} & $0.25\times$ & 85.52 & 84.96 & 85.43 & \textbf{85.64} & 78.49 & 78.53 & 79.57 & \textbf{80.83} & 54.68 & 52.06 & 47.82 & \textbf{79.35} \\
  & $0.5\times$ & 87.26 & 89.58 & 88.78 & \textbf{89.63} & 85.21 & 85.41 & 85.62 & \textbf{87.12} & 79.60 & 76.98 & 74.25 & \textbf{86.63} \\
  & $0.75\times$ & \textbf{91.28} & 91.04 & 90.06 & 91.14 & 88.71 & 88.76 & 87.97 & \textbf{89.33} & 81.22 & 72.71 & 68.26 & \textbf{86.03} \\
  & $1.0\times$ & 92.02 & 91.52 & 90.32 & \textbf{92.06} & 87.72 & 87.90 & 88.37 & \textbf{89.26} & 82.46 & 77.75 & 75.24 & \textbf{86.90} \\
  & Avg. & 89.02 & 89.28 & 88.65 & \textbf{89.62} & 85.03 & 85.15 & 85.38 & \textbf{86.64} & 74.49 & 69.88 & 66.39 & \textbf{84.73} \\
 \hline
 \specialrule{0em}{0pt}{2.5pt}
 \multirow{5}{*}{CIFAR-100} & $0.25\times$ & 26.88 & 27.01 & \textbf{27.25} & 27.06 & 8.09 & 7.01 & 12.17 & \textbf{20.53} & 7.41 & 8.10 & 10.13 & \textbf{19.40} \\
  & $0.5\times$ & 37.02 & 37.61 & 37.55 & \textbf{37.96} & 16.75 & 17.43 & 18.27 & \textbf{31.80} & 16.50 & 15.73 & 16.87 & \textbf{29.46} \\
  & $0.75\times$ & 43.27 & 43.55 & 43.76 & \textbf{43.94} & 23.63 & 27.22 & 25.08 & \textbf{39.66} & 27.81 & 25.64 & 23.76 & \textbf{39.37} \\
  & $1.0\times$ & 47.61 & 47.64 & 47.75 & \textbf{47.84} & 29.10 & 28.63 & 30.14 & \textbf{40.01} & 30.36 & 32.64 & 28.46 & \textbf{39.72} \\
  & Avg. & 38.70 & 38.95 & 39.08 & \textbf{39.20} & 19.39 & 20.07 & 21.42 & \textbf{33.00} & 20.52 & 20.53 & 19.81 & \textbf{31.99} \\
 \hline
 \hline
\end{tabular}}
\end{center}
\caption{Experimental results on Benchmark Comparison Evaluation. We report classification performance using the standard Top-1 accuracy (\%) as the evaluation metric. The column labeled "Avg." denotes the average accuracy across all model scales, providing a comprehensive view of overall performance. Results highlighted in "\textbf{numbers}" indicate the best performance achieved within each comparison group.}
\label{tab1:benmarkexp}
\end{table*}

In the above heterogeneous FL framework, the global $\overline{B}$ measures information across all model scales, and the global list $\{\overline{M}^{\text{p}}\}$ contains heterogeneous information for each model scale. 
In heterogeneous settings, each model scale is typically associated with a limited number of clients and data, which can lead to severe information loss, particularly detrimental for SNNs with inherently sparse feature representations. 
On the other hand, the spike firing rate in SNNs serves as a distinctive property that effectively reflects the information density received by each network layer.

Thus, this section proposes a novel heterogeneous information fusion module based on firing rates, called the HIFI module. 
The key idea of HIFI is to identify the most informative layer parameters based on their firing rates and fuse these heterogeneous weights through information decoupling. 
This approach enriches the information representation of global SNNs and further enhances overall model performance. 
Firstly, we define the information density of layer $L$ via the following layer spike firing rate $r_L$,
\begin{equation}
    r_L = \frac{1}{\mathcal{T}\cdot||~L~||}\sum\limits_{t=1}^\mathcal{T}\sum\limits_{l \in \text{layer}~L}S_l[t],
\end{equation}
where $l$ means the spike neuron in Layer $L$, and $||~L~||$ denotes the number of all spike neurons in Layer $L$, $S_l[t]$ is the output of neuron $l$ at time-step $t$, $\mathcal{T}$ is the total time window length. 

Then, we compare the firing rates $r_L$ across all model scales in \textbf{p} to identify the layer with the highest information density $\hat{L}$, which is subsequently used for the heterogeneous information fusion,
\begin{align}
    \hat{L}&=\text{mode}(\{L^{p_i}|p_i\in\textbf{p}\}), \\
    L^{p_i}&=\underset{L}{\text{argmax}}(\{r_L^{p_i}|L\in \text{layer}~1,2,...\}),
\end{align}
where $\text{mode}(\cdot)$ means the mode extraction, $r_L^{p_i}$ denotes the spike firing rate of layer $L$ with $p_i \times$ width.
After identifying the layer with the highest information density, we perform heterogeneous information decoupling on the $\hat{L}$ layer of each model scale, following the Tucker decomposition framework~\cite{sec3_tucker2009tensor,sec3_tucker2019},
\begin{align}
    \overline{M}^{p_i}_{\hat{L}} \approx G^{p_i} \times_1 A^{p_i}_1 \times_2 A^{p_i}_2 \times_3 A^{p_i}_3 \times_4 A^{p_i}_4,
\end{align}
where $\overline{M}^{p_i}_{\hat{L}}$ is the model scale factor of layer $L$ in the global $p_i\times$ model. 
$G^{p_i} \in \mathbb{R}^{b_1 \times b_2 \times b_3^2}$ is the core tensor, and it retains the same shape when decomposed from all model scale factors $\{\overline{M}^{\mathbf{p}}_{\hat{L}}\}$.
$b_1, b_2$ and $b_3$ are pre-defined Tucker ranks. 
$[A^{p_i}_1, A^{p_i}_2, A^{p_i}_3, A^{p_i}_4]$ is the list of two-dimensional factor matrices, and $\times_k$ is the Mode-$k$ Product for tensor and matrix.
Finally, we aggregate the core tensors to get $\overline{G}$ and reconstruct $\{\overline{M}^{\mathbf{p}}_{\hat{L}}\}$ using the factor matrices,
\begin{equation}
    \overline{G} = \frac{1}{|~\mathbf{p}~|}\sum\limits_{p_i \in \mathbf{p}} G^{p_i} \in \mathbb{R}^{b_1 \times b_2 \times b_3^2},
\end{equation}
\begin{equation}
    \overline{M}^{p_i}_{\hat{L}} \leftarrow \overline{G} \times_1 A^{p_i}_1 \times_2 A^{p_i}_2 \times_3 A^{p_i}_3 \times_4 A^{p_i}_4.
\end{equation}
In summary, to mitigate the susceptibility of SNNs to information loss caused by their inherent sparse representations, the SFedHIFI framework identifies high-information-density layers via their spiking firing rates, then performs heterogeneous information fusion of global model scale factors $\{\overline{M}^\textbf{p}\}$ through information decoupling, thereby further enhancing the performance of the global model.

\subsection{The Local Objective Function}
Our SFedHIFI employs the temporal efficient training (TET) loss~\cite{sec3_tetloss}, specifically adapted for SNNs, as the primary optimization objective for local training. 
As shown in Eq.\ref{eq:mdec4}, the training process simultaneously optimizes both the parameter matrices $B$ and $M^\textbf{p}$. 
To enhance the global generalization capacity, the shared basis $B$ needs to encode comprehensive knowledge representations. However, the standard objective function fails to effectively constrain the training dynamics of $B$ to achieve this goal. To this end, we introduce an additional orthogonal regularizer into the local loss function, following~\cite{sec1_HFL_rlanc},
\begin{equation}
    \mathcal{L} = \mathcal{L}_{tet}\left(\mathbf{y}, y\right)+\lambda \cdot \sum^{L}\Vert B_L \cdot B_L^{T} - W_\text{I} \Vert_2,
\end{equation}
\begin{equation}
    \mathcal{L}_{tet} = \frac{1}{\mathcal{T}} \sum_{t=1}^{\mathcal{T}} \mathcal{L}_{CE}\left(\mathbf{y}[t], y\right),
\end{equation}
where $\lambda$ is the balancing factor, and $W_\text{I}$ means identity matrix. $\mathbf{y}[t]$ denotes the predicted output of SNN at time step $t$, $y$ is the label. $B_L$ is the shared basis of layer $L$, $B_L^T$ is the transpose of $B_L$. 
In this work, we impose an orthogonality constraint on $B$ via $L_2$ normalization, which implicitly enriches its knowledge representation capacity.

\section{Experiments}
This section provides a comprehensive evaluation of the proposed SFedHIFI framework. We first benchmark it against several classical FL methods with SNN, then perform ablation studies to assess the contribution of HIFI module. Finally, we compare SFedHIFI with ANN-based heterogeneous FL in terms of performance and energy efficiency.
The source code and Appendix are available at https://github.com/rtao499/SFedHIFI.

\subsection{Experimental Setup}

\textbf{Datasets}.
We conducted experiments using three popular datasets, Fashion-MNIST~\cite{sec4_fmnist}, CIFAR-10, and CIFAR-100~\cite{sec4_cifar10_cifar100}. 
And like previous works~\cite{sec2_HFL_moon,sec4_niid}, we partition the datasets across clients under both IID and Non-IID data settings. 
The specifications of the dataset and Data Partitioning are detailed in the Appendix.

\textbf{Model Selection}. 
In terms of model structure, we adopt an identical SNN with the same setting for each dataset to ensure fair comparison. For Fashion-MNIST, we choose a simple 4-layer SNN (64C3-128C3-MP2-128C3-MP2-FC). And we use Spiking-ResNet18 provided by Spikejelly \citep{sec4_spikingjelly} for CIFAR-10 and CIFAR-100. For the SNN implementation, we set LIF neurons with threshold voltage $V_{\text{th}}=1.0$, and the total time step $\mathcal{T}$ is set to $10$.

\textbf{Heterogeneity Settings}. 
In all experiments, we assume a total of 10 clients, with 50\% of them randomly selected to participate in each communication round of SFedHIFI. To simulate system heterogeneity in model capacity, we set four distinct network scales, represented as $\textbf{p} = \{0.25, 0.5, 0.75, 1.0\}$. For local data heterogeneity, we follow a Dirichlet-based partitioning strategy, setting the concentration parameter $\alpha$ to 0.5 and 0.3, respectively, to reflect varying degrees of non-IID data distributions across clients.

\textbf{Baseline Approaches and Implementation Details}.
For a comprehensive comparison, we compare our SFedHIFI with three well-established FL baselines, including: FedAVG~\cite{sec3_fedavg}, FedProx~\cite{sec2_HFL_fedprox}, and FedNova~\cite{sec4_fednova}. 
Due to model heterogeneity, the baseline methods perform aggregation only among models of the same scale during the FL process.
Further details of baseline and implementation are available in the Appendix.
\begin{table*}[htbp]
\begin{center}
\resizebox{1.0\linewidth}{!}{
\begin{tabular}{ccccccccccccc}
\hline
\hline
\specialrule{0em}{0pt}{3pt}
\multirow{2}{*}{\textbf{Scale}} & \multicolumn{4}{c}{\textbf{IID}} & \multicolumn{4}{c}{\textbf{Non-IID} Dir(0.5)} & \multicolumn{4}{c}{\textbf{Non-IID} Dir(0.3)} \\
 \cmidrule(lr){2-5} \cmidrule(lr){6-9} \cmidrule(lr){10-13}
 & $\text{HIFI}_{w/o}$ & $\text{HIFI}_{\text{fixed}}$ & $\text{HIFI}_{\text{rand}}$ & \textbf{SFedHIFI} & $\text{HIFI}_{w/o}$ & $\text{HIFI}_{\text{fixed}}$ & $\text{HIFI}_{\text{rand}}$ & \textbf{SFedHIFI} & $\text{HIFI}_{w/o}$ & $\text{HIFI}_{\text{fixed}}$ & $\text{HIFI}_{\text{rand}}$ & \textbf{SFedHIFI} \\
 \hline
 \specialrule{0em}{0pt}{2pt}
 $0.25\times$ & 83.82 & 82.11 & 77.72 & \textbf{85.64} & 80.65 & 74.25 & 73.60 & \textbf{80.83} & 74.52 & 68.94 & 58.57 & \textbf{79.35} \\
 $0.5\times$ & 88.83 & 89.07 & 88.36 & \textbf{89.63} & 84.87 & 85.90 & 83.56 & \textbf{87.12} & 76.16 & 82.20 & 72.70 & \textbf{86.63} \\
 $0.75\times$ & 90.55 & 91.06 & 89.96 & \textbf{91.14} & 85.55 & 88.94 & 86.61 & \textbf{89.33} & 82.68 & 82.26 & 79.19 & \textbf{86.03} \\
 $1.0\times$ & 91.62 & 91.81 & 91.15 & \textbf{92.06} & 85.42 & 89.08 & 85.98 & \textbf{89.26} & 79.29 & 84.39 & 59.46 & \textbf{86.90} \\
 Avg. & 88.71 & 88.52 & 86.80 & \textbf{89.62} & 84.12 & 84.54 & 82.44 & \textbf{86.64} & 78.16 & 79.45 & 67.48 & \textbf{84.73} \\
 \hline
 \specialrule{0em}{0pt}{2pt}
 \hline
\end{tabular}}
\end{center}
\caption{Ablation experiment results of the HIFI module on CIFAR-10. We still report classification performance using the standard Top-1 accuracy (\%) as the evaluation metric.}
\label{tab2:ablationexp}
\end{table*}

\subsection{Benchmark Comparison Evaluation}
In the benchmark comparison experiments, we demonstrate the advantages of SFedHIFI by comparing against three baselines of FedAVG, FedProx, and FedNova adapted to SFL across three standard datasets. 
Specifically, we simulate an SFL environment with 10 clients, each assigned a local dataset following either an IID or non-IID distribution. 
In each communication round, 5 active clients are randomly selected to participate, each equipped with a heterogeneous SNN model. 
Moreover, we define four different SNN model scales based on network width to build model heterogeneity. 
While SFedHIFI performs cross-scale aggregation across all four model types, the baseline methods apply their respective aggregation strategies only among models of the same scale.

Table~\ref{tab1:benmarkexp} shows that the proposed SFedHIFI consistently outperforms all baselines across all model scales, datasets, and data distributions. On the relatively simple Fashion-MNIST dataset, the performance gap among all methods is within 2.5\% accuracy. Under the IID setting, FedAVG achieves the best performance among the baselines. In contrast, SFedHIFI benefits from the fusion of heterogeneous information across all model scales, allowing it to learn a broader knowledge base and achieve the best results at every model scale, with an average accuracy improvement of 1.02\%. As the degree of non-IID increases, although SFedHIFI still maintains superior overall performance, FedProx and FedNova begin to demonstrate their advantages in handling data heterogeneity, outperforming FedAVG and performing comparably to SFedHIFI. These results not only validate the effectiveness of SFedHIFI in heterogeneous information fusion but also indicate that the simplicity of Fashion-MNIST may limit its ability to highlight the differences among methods.

On more complex datasets, SFedHIFI demonstrates more significant improvements. On CIFAR-10, the accuracy gains can reach up to 30\%. Under the IID setting, all four methods perform similarly, as models of all scales have equal access to the same data knowledge. Nevertheless, SFedHIFI still achieves the best overall performance, showcasing the advantage of heterogeneous information fusion. As the degree of non-IID increases, the superiority of our method becomes more pronounced, especially under $Dir(0.3)$. In this case, clients contain highly diverse data knowledge, and aggregating within homogeneous model structures not only limits knowledge sharing but also exacerbates performance degradation for models with a smaller scale. In contrast, SFedHIFI significantly improves the performance of small-scale models compared to baseline methods, confirming its effectiveness. These results also validate that cross-scale aggregation can successfully fuse heterogeneous information, enabling compact models to acquire deeper and broader knowledge.

On CIFAR-100, none of the methods achieved high absolute performance due to the dataset’s inherent complexity. However, this complexity also provides a clearer differentiation among methods. Similar to previous datasets, SFedHIFI shows a slight advantage under the IID setting. 
Under non-IID settings, baseline models at all scales struggle to learn effectively from local data. This highlights the inconsistency of data knowledge across clients, where aggregating only within the same model scale fails to build a comprehensive knowledge base. Consequently, large-scale models suffer from incomplete learning, while the limited capacity of small-scale models is further exposed without the guidance of larger models. 
In contrast, SFedHIFI consistently improves models of all scales, outperforming baselines by up to 19\% in accuracy. These results demonstrate that, even on complex datasets, SFedHIFI can effectively expand the knowledge coverage of all clients by fusing heterogeneous information across models of different scales.

\subsection{Ablation Experiment of the HIFI Module}
In the proposed SFedHIFI framework, we design a tailored HIFI Module for SNNs. To evaluate its effectiveness, we conduct experiments on the CIFAR-10 dataset. Specifically, the HIFI Module identifies high-information-density layers based on spike firing rates and performs heterogeneous information fusion across different model scales. To analyze the impact of this design, we conduct three comparative experiments: (i) $\text{HIFI}_{w/o}$ – without heterogeneous information fusion; (ii) $\text{HIFI}_{\text{fixed}}$ – information fusion is performed on a fixed layer; and (iii) $\text{HIFI}_{\text{rand}}$ – information fusion is performed on randomly selected layers. The experimental results are presented in Table~\ref{tab2:ablationexp}.

From ablation results in Table~\ref{tab2:ablationexp}, our proposed method achieves the best performance across all comparisons, with particularly significant gains under the Dir(0.3) setting. Specifically, in all scenarios, $\text{HIFI}_{\text{rand}}$ performs the worst, indicating that the informativeness and error sensitivity of SNN layers vary significantly. Randomly selecting layers for aggregation may even introduce harmful effects. 

When comparing SFedHIFI with $\text{HIFI}_{w/o}$, we observe that under IID settings, the HIFI module brings more substantial improvements to smaller models. This is because, with uniform data distributions, large models gain little from small models in terms of their local data knowledge, whereas small models benefit from the parameter guidance of larger models. Under non-IID settings, SFedHIFI provides significant improvements across all model scales, confirming the effectiveness of the HIFI module. These results also validate that due to the sparsity of SNN representations and the information loss induced by pruning, removing the HIFI module results in limited aggregation effectiveness.

In the comparison between SFedHIFI and $\text{HIFI}_{\text{fixed}}$, SFedHIFI yields greater improvements for smaller models. 
It is worth noting that the fixed layer in $\text{HIFI}_{\text{fixed}}$ is set to the last convolutional layer of Spiking-ResNet18, which is empirically regarded as one of the most informative layers in conventional ResNet18 architectures.
However, our results suggest that information distribution in SNNs does not entirely align with ANN-based assumptions. By leveraging spike firing rates, our approach enables a more refined identification of informative layers, thereby capturing richer and more effective heterogeneous information for aggregation.

\begin{table}[htbp]
\begin{center}
\resizebox{1.0\linewidth}{!}{
\begin{tabular}{cccccccccc}
\hline
\hline
\specialrule{0em}{0pt}{3pt}
\multirow{2}{*}{\textbf{Metric}} & \multirow{2}{*}{\textbf{Scale}} & \multicolumn{2}{c}{\textbf{IID}} & \multicolumn{2}{c}{\textbf{Dir(0.5)}} & \multicolumn{2}{c}{\textbf{Dir(0.3)}} \\
 \cmidrule(lr){3-4} \cmidrule(lr){5-6} \cmidrule(lr){7-8}
 & & ANNs & \textbf{Ours} & ANNs & \textbf{Ours} & ANNs & \textbf{Ours} \\
 \hline
 \specialrule{0em}{0pt}{2pt}
 \multirow{5}{*}{\shortstack[c]{\textbf{Accuracy} \\ (\%)~$\uparrow$}} & $0.25\times$ & 89.75 & \textbf{91.46} & \textbf{90.91} & 88.70 & \textbf{90.62} & 87.27 \\
 & $0.5\times$ & 90.14 & \textbf{92.38} & \textbf{91.74} & 89.85 & \textbf{91.00} & 88.31 \\
 & $0.75\times$ & 90.90 & \textbf{92.03} & \textbf{91.33} & 90.41 & \textbf{91.04} & 88.99 \\
 & $1.0\times$ & 91.01 & \textbf{92.43} & \textbf{91.90} & 90.32 & \textbf{91.56} & 89.18 \\
 & Avg. & 90.45 & \textbf{92.08} & \textbf{91.47} & 89.82 & \textbf{91.06} & 88.43 \\
 \specialrule{0em}{0pt}{2pt}
 \hline
 \hline
 \specialrule{0em}{0pt}{2pt}
 & & \multicolumn{3}{c}{ANNs} &\multicolumn{3}{c}{\textbf{Ours}} \\
 \hline
 \specialrule{0em}{0pt}{2pt}
 \multirow{5}{*}{\shortstack[c]{\textbf{Energy} \\ ($\times 10^{-3}$mj)~$\downarrow$}} & $0.25\times$ & \multicolumn{3}{c}{34.059} & \multicolumn{3}{c}{\textbf{23.181}} \\
 & $0.5\times$ & \multicolumn{3}{c}{69.503} & \multicolumn{3}{c}{\textbf{43.354}} \\
 & $0.75\times$ & \multicolumn{3}{c}{108.639} & \multicolumn{3}{c}{\textbf{62.320}} \\
 & $1.0\times$ & \multicolumn{3}{c}{151.469} & \multicolumn{3}{c}{\textbf{75.737}} \\
 & Avg. & \multicolumn{3}{c}{90.917} & \multicolumn{3}{c}{\textbf{51.148}} \\
 \specialrule{0em}{0pt}{2pt}
 \hline
 \hline
\end{tabular}}
\end{center}
\caption{Comparison between SFedHIFI and ANN in Accuracy and Energy Consumption under Identical Network Configurations on Fashion-MNIST.}
\label{tab3:eng}
\end{table}

\subsection{Energy Analysis}
SNNs have been widely recognized as a promising low-power alternative to ANNs. Therefore, we are particularly interested in comparing the advantages of ANN-based FL and SFL. To enable a more intuitive comparison between SNNs and ANNs, we evaluate both model performance and energy consumption on the Fashion-MNIST dataset. Specifically, we replace the SNN models in SFedHIFI with ANNs of the same scale and remove the Heterogeneous Information Fusion module to construct a corresponding ANN-based FL system.

In terms of model performance, we compare the accuracy of global models at four different network scales. For energy consumption, we follow prior work~\cite{sec4_snncomp} by estimating the synaptic operations (SynOPs) of SNNs using the accumulated number of multiply-and-accumulate (MAC) and spike-based accumulate (AC) operations. For ANNs, we compute energy consumption based on floating-point operations (FLOPs), following the method in~\cite{sec4_flops}. We provide the energy estimation formulas used for both SNN and ANN models below:
\begin{equation}
\label{eq15:EANN}
\begin{split}
E_{\text{ANN}}&= \sum_{l=1}^{L} (\text{FLOPs}^{(l)} \times E_{\text{MAC}}) \\
&=\sum_{l=1}^{L} 2(f_{\text{in}}^l+1)\beta^{l} \times E_{\text{MAC}},
\end{split}
\end{equation}
\begin{equation}
\label{eq16:ESNN}
\begin{split}
E_{\text{SNN}}&=(\text{FLOPs}^{(1)} \times E_{\text{MAC}})+
\sum_{l=2}^{L} (\text{SynOPs}^{(l)} \times E_{\text{AC}}) \\
&=(\text{FLOPs}^{(1)} \times E_{\text{MAC}})+\\&~~~~~~~~~~~~~~~~~~~~\sum_{l=2}^{L}\text{FLOPs}^{(l)} \times \overline{r}^{(l)} \times \mathcal{T} \times E_{\text{AC}},
\end{split}
\end{equation}
where $L$ is the total number of layers in both ANN and SNNs, and $l$ is the layer index, $\text{FLOPs}^{(l)}/\text{SynOPs}^{(l)}$ is FLOPs/SynOPs in the $l$-th layer. $f_{\text{in}}^l$ denotes the number of incoming connections to a neuron of the $l$-th layer. $\beta^l$ represents the number of neurons in the $l$-th layer.
$\overline{r}^{(l)}$ means the average firing rate per time step of the neurons in the $l$-th layer.
$E_{\text{MAC}}$ is the energy cost per MAC operation, and $E_{\text{AC}}$ is the energy cost per AC operation. We set $E_{\text{MAC}}=4.6\text{pJ}$ and $E_{\text{AC}}=0.9\text{pJ}$, following~\cite{sec4_spikformer}.

For ANNs, we calculate the FLOPs for each layer and multiply it by $E_{\text{MAC}}$ to estimate the energy consumption. 
For SNNs, when processing static RGB image data, the first layer is responsible for encoding the raw input into spike sequences. As a result, the operations in the first layer are computed as standard $\text{FLOPs}^{(1)}$. The following layers, which process binary spike inputs, rely on synaptic operations for computation.
We report the accuracy and the energy consumption for each prediction in different-scale ANNs and SNNs in Table \ref{tab3:eng}.

As shown in Table~\ref{tab3:eng}, SNN models across all scales consistently exhibit lower energy consumption compared to their ANN counterparts. To provide a comprehensive evaluation, we first average the "Avg." Top-1 accuracy across the three partitioning settings (IID, Dir(0.5), and Dir(0.3)), and then compute the relative accuracy drop of SNNs compared to ANNs, alongside the average energy consumption. Overall, SNNs achieve a 43.74\% reduction in power consumption with less than 1\% accuracy degradation, which is an acceptable trade-off for energy-efficient learning.
Interestingly, under the IID setting, our SFedHIFI even outperforms the ANN-based method in terms of accuracy. We attribute this to the advantage of the HIFI module in effectively leveraging heterogeneous information. 
However, as the Non-IID degree increases, the robustness advantage brought by the higher precision of ANNs becomes increasingly evident.
Moreover, we observe that the benefits of SNNs are more pronounced in larger models: as network width increases, the energy savings of SNNs become more significant, while accuracy loss remains marginal. This energy-accuracy trade-off highlights the unique strength of SNNs in resource-constrained edge scenarios, demonstrating the practical potential of SFedHIFI in real-world SFL deployments.

\section{Conclusion}
In this work, we propose SFedHIFI, a novel heterogeneous SFL framework. Based on Channel-wise Matrix Decomposition, SFedHIFI enables clients to adaptively deploy SNN models with varying widths according to their local resource constraints. 
To address the severe information loss caused by the inherent sparsity of SNNs, we design a Firing Rate-Based HIFI Module. 
Leveraging spiking dynamics, this module identifies high-information-density layers based on spike firing rates and performs Tucker information decoupling to fuse valuable heterogeneous layers from the model scale factors.
Experimental results across multiple benchmarks demonstrate that SFedHIFI consistently outperforms all baseline SFL methods, effectively enabling heterogeneous federated learning via channel pruning. Ablation studies confirm the effectiveness of the proposed HIFI module and validate the mapping between firing rates and spiking information density. Furthermore, energy consumption analysis shows that SFedHIFI achieves notable energy savings with minimal accuracy trade-off.
Overall, this work demonstrates the feasibility and effectiveness of constructing heterogeneous SFL systems, laying the foundation for broader deployment of neuromorphic intelligence in real-world federated learning scenarios.

\section{Acknowledgments}
This research was supported by the NSFC (62376228), the Fundamental Research Funds for the Central Universities (JBK2406079), and the Graduate Representative Achievement
Cultivation Project of Southwest University of Finance and Economics (JGS2024067).

\bibliography{aaai2026}

\end{document}